\pdfoutput=1

\documentclass[11pt]{article}

\usepackage{acl2023}

\usepackage{times}
\usepackage{latexsym}

\usepackage[T1]{fontenc}

\usepackage[utf8]{inputenc}

\usepackage{microtype}
\usepackage{xspace,mfirstuc,tabulary}
\usepackage{enumitem}
\setitemize{noitemsep,topsep=0pt,parsep=0pt,partopsep=0pt,leftmargin=*}
\usepackage{amsthm}
\usepackage{amsmath}
\usepackage{mathtools}
\usepackage{amssymb}
\usepackage{multirow}
\usepackage{comment}
\usepackage{array}
\usepackage{tabularx}
\definecolor{darkspringgreen}{rgb}{0.09, 0.45, 0.27}
\usepackage{graphicx}
\usepackage[singlelinecheck=false,justification=justified]{caption}
\usepackage{subcaption}

\usepackage{scalerel,stackengine}
\stackMath
\newcommand\reallywidehat[1]{%
\savestack{\tmpbox}{\stretchto{%
  \scaleto{%
    \scalerel*[\widthof{\ensuremath{#1}}]{\kern-.6pt\bigwedge\kern-.6pt}%
    {\rule[-\textheight/2]{1ex}{\textheight}}
  }{\textheight}%
}{0.5ex}}%
\stackon[1pt]{#1}{\tmpbox}%
}
\makeatletter
\renewcommand\@makefntext[1]{\leftskip=2em\hskip-2em\@makefnmark#1}
\makeatother

\usepackage{algorithm,algpseudocode}
\newlength{\bibitemsep}
\newlength{\bibparskip}
\let\oldthebibliography\thebibliography
\renewcommand\thebibliography[1]{%
  \oldthebibliography{#1}%
  \setlength{\parskip}{\bibitemsep}%
  \setlength{\itemsep}{\bibparskip}%
}

\usepackage{enumitem}
\setlist{nolistsep}
\theoremstyle{definition}

\usepackage[]{xcolor}

\DeclareMathOperator*{\argmin}{arg\,min}

\newcolumntype{P}[1]{>{\centering\arraybackslash}p{#1}}

\newcommand{\Pruthi}{\textsc{CharAtt}\xspace}
\newcommand{\Alzantot}{\textsc{WordAtt}\xspace}

\newcommand{\NNIF}{\textsc{NNIF}\xspace}
\newcommand{\FastIF}{\textsc{FastIF}\xspace}
\newcommand{\Mahalanobis}{\textsc{Mahal}\xspace}
\newcommand{\RSV}{RSV\xspace}
\newcommand{\SHAP}{SHAP\xspace}
\newcommand{\FGWS}{FGWS\xspace}
\newcommand{\DISP}{DISP\xspace}
\newcommand{\LID}{LID\xspace}
\newcommand{\MDRE}{MDRE\xspace}
\newcommand{\IMDB}{IMDB\xspace}
\newcommand{\Mnli}{\textsc{MultiNLI}\xspace}
\newcommand{\tsne}{t-SNE\xspace}

\newcommand{\knn}{\textsc{kNN}\xspace}
\newcommand{\dknn}{\textsc{DkNN}\xspace}
\newcommand{\zt}{\ensuremath{z_\textsubscript{test}}}
\newcommand{\ra}{\ensuremath{\mathcal{R}}}
\newcommand{\di}{\ensuremath{\mathcal{D}}}

\definecolor{mycolor}{RGB}{202, 42, 219}

\newcommand{\myterm}[1]{\textsc{#1}}
\newcommand{\mypar}[1]{\noindent\textbf{#1}}

\usepackage{booktabs}
\usepackage[justification=centering]{caption}

\usepackage{supertabular}
\usepackage{adjustbox}

\title{What Learned Representations and Influence Functions Can Tell Us About Adversarial Examples}

\author{Shakila Mahjabin Tonni \and Mark Dras \\
School of Computing, Macquarie University \\  \texttt{shakila.tonni@mq.edu.au, mark.dras@mq.edu.au}}

\begin{document}
\maketitle
\begin{abstract}

Adversarial examples, deliberately crafted using small perturbations to fool deep neural networks, were first studied in image processing and more recently in NLP.  While approaches to detecting adversarial examples in NLP have largely relied on search over input perturbations, image processing has seen a range of techniques that aim to characterise adversarial subspaces over the learned representations.  

In this paper, we adapt two such approaches to NLP, one based on nearest neighbors and influence functions and one on Mahalanobis distances.  The former in particular produces a state-of-the-art detector when compared against several strong baselines; moreover, the novel use of influence functions provides insight into how the nature of adversarial example subspaces in NLP relate to those in image processing, and also how they differ depending on the kind of NLP task. 
\end{abstract}

\section{Introduction}
\label{sec:Introduction}
The high sensitivity of deep neural networks (DNNs) to slight modifications of inputs is widely recognised and makes DNNs a convenient target for adversarial attacks \citep{szegedy2013intriguing}. Creating malicious inputs or adversarial examples by adding small perturbations to the model’s inputs can cause the model to misclassify the inputs that would be predicted correctly otherwise. Such adversarial attacks are highly successful in both image and Natural Language Processing (NLP) domains. 

In the image domain, due to the straightforwardness of creating adversarial images by calibrating noise to the original records, researchers have explored many high-performing adversarial attacks~\citep[for example]{%
papernot2016limitations, 
moosavi2016deepfool, carlini2017towards%
}. The perturbations of the input images degrade the model's performance with a high success rate and are generally imperceptible to a human. 

Work in the NLP space has followed that in image processing.  Here,
in addition to the goal of impacting the model’s prediction, adversarial text examples need to be syntactically and semantically sound to the reader. Consequently, adversarial attack techniques on text use semantics-preserving textual changes at the character level, word level and phrase level or sentence level \citep[for example]{pruthi-etal-2019-combating,
alzantot-etal-2018-generating, 
li-etal-2020-bert-attack%
}. 
Table~\ref{tab:text_adv} illustrates two examples, showing different types of attack formulation in NLP.

In the image domain, defence against adversarial attack can be `proactive' or `reactive' \citep{CVPRPaper}, where proactive defence refers to improving the model’s robustness \citep{madry2017towards,Gopinath2018robust,cohen2019certified} and reactive defence focuses on detecting real adversarial examples before they are passed to neural networks \citep{feinman2017detecting, ma2018characterizing, lee2018simple, papernot2018deep}. Broadly speaking, for reactive methods, the detection of adversarial examples involves taking a conceptualisation of the space of learned representations and the adversarial subspaces within them \citep{tanay2016boundary,tramer2017space}, and then characterising the differences in some function of the learned representations between the actual and the adversarial inputs produced by the DNN; for example, \citet{ma2018characterizing} applied a local intrinsic dimensionality (\LID) measure to the learned representations and used that to successfully distinguish normal and adversarial images.

In the NLP space, relatively fewer adversarial defence techniques have been proposed. Among them, many focus on enhancing the models’ robustness proactively through adversarial training~\citep{jia-etal-2019-certified, pruthi-etal-2019-combating, jin2020bert}; generating textual samples for proactive adversarial training is computationally expensive because of necessary search and constraints based on sentence encoding \citep{yoo-qi-2021-towards-improving}. 
Reactive adversarial text detection techniques have mostly been different from their image counterparts, in that they typically modify the input by e.g. 
repeatedly checking word substitutions~\citep{mozes-etal-2021-frequency, wang2022detecting, zhou-etal-2019-learning} rather than trying to characterise the learned representations; consequently, they focus on detecting synonym-substitution adversarial examples. An exception is the work of \citet{liu-etal-2022-detecting}, which both adapts \LID to the text space and proposes the new MultiDistance Representation Ensemble (\MDRE) method; their state-of-the-art results suggest that the detection methods based on learned representations drawn from the image processing domain are a promising source of ideas for NLP.

\begin{figure}
    \captionsetup{singlelinecheck = false, justification=justified}
    \includegraphics[width=0.45\textwidth]{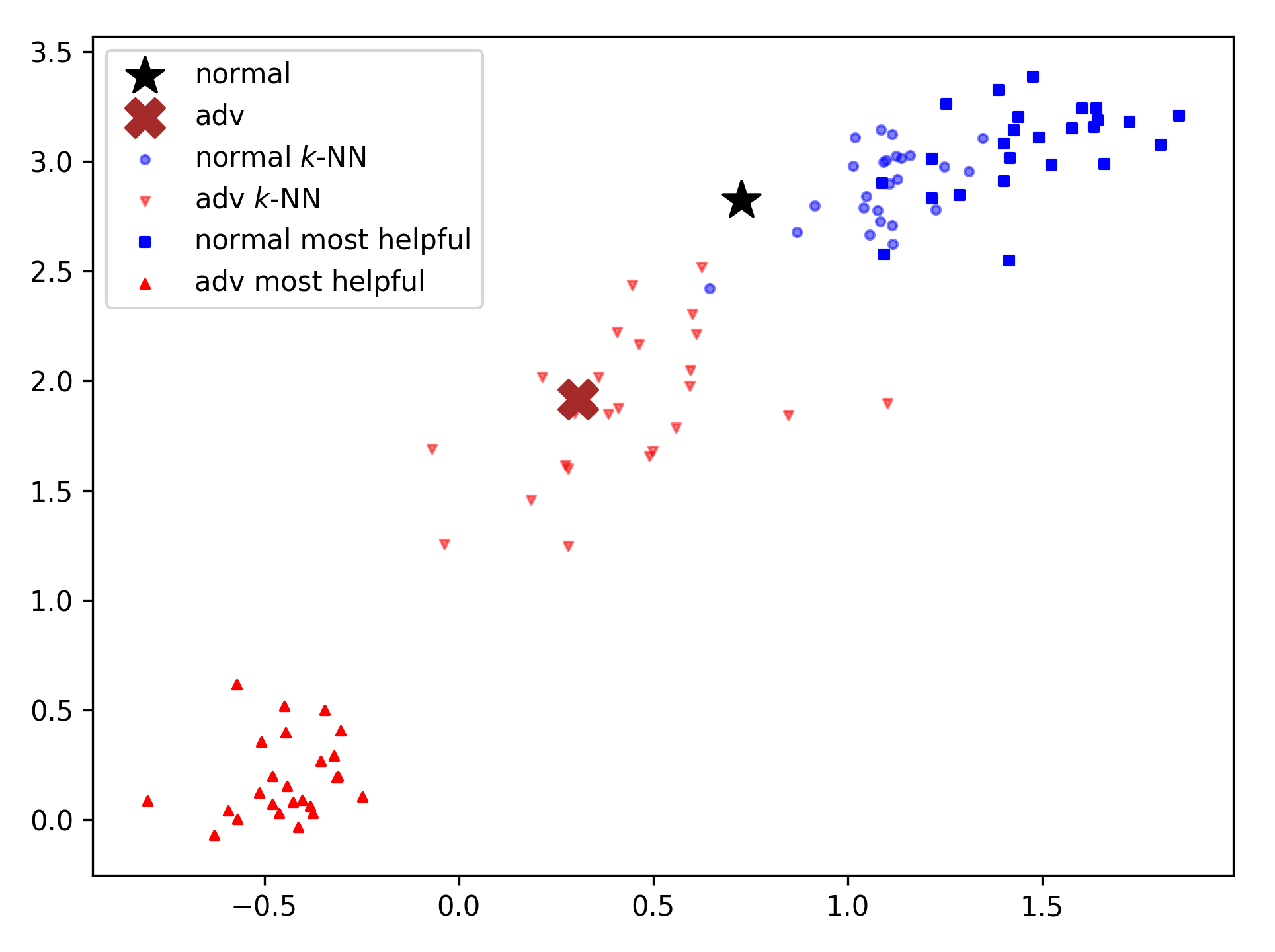}
    \caption{Adversarial examples characterised by divergence in learned representations between nearest neighbors and training points selected by influence functions, unlike original examples (from~\citep{CVPRPaper}).}
    \label{fig:cohen_embed_space}
    \vspace{-1em}
\end{figure}

The particular focus of the present paper is the use of influence functions in adversarial detection methods, proposed for image processing by \citet{CVPRPaper}. They propose that distances to nearest neighbors (used by previous methods) and influence functions, which measure the impact of every training sample on validation or test set data, can be used complementarily to detect adversarial examples: they argue, with support from the strong results from their method, that adversarial examples locate in different regions of the learned representation space of their neighbors with respect to influence functions, compared to original datapoints (Fig~\ref{fig:cohen_embed_space}). Specifically, in the image space, for original datapoints, nearest neighbors and influence function training points overlap, but for adversarial examples, they do not. Influence functions have only relatively recently begun to be explored in NLP, with \citet{han-etal-2020-explaining} finding that, with the variety of classification tasks in NLP, the information provided by influence functions differs from image processing and is task-dependent. In this paper, noting significant differences between inputs in NLP and image processing (continuous versus discrete) and attack types, we explore whether and how they can help in NLP in detecting adversarial examples using learned representations, and what this can tell us about the nature of adversarial subspaces.

We also adapt a second method from the image processing literature, by \citet{lee2018simple}, which uses a Mahalanobis-based confidence score; this was a strong baseline for \citet{CVPRPaper}, 
giving an additional perspective on the nature of adversarial subspaces in NLP.
  
The contributions of this paper are as follows:
\begin{itemize}
    \item An adaptation of two adversarial detection techniques from the image processing literature, \Mahalanobis confidence~\citep{lee2018simple} and Nearest Neighbor Influence Functions (NNIF)~\citep{CVPRPaper}, into the text domain; we show that we can achieve SOTA results relative to several strong, recent baselines. 

    \item An analysis of how influence functions work in this context, contributes to understanding both the nature of adversarial subspaces in the text space and what information influence functions can provide.

\end{itemize}

\begin{table*}
\centering
\small
\begin{tabular}{@{}lll@{}}
\toprule
Original Text & \begin{tabular}[c]{@{}l@{}}at last, a movie that handles the probability of alien visits with \\the appropriate depth and loving warmth. \end{tabular}
& Positive \\ \midrule

\begin{tabular}[c]{@{}l@{}}Char-level\\\cite{pruthi-etal-2019-combating} \end{tabular}  & \begin{tabular}[c]{@{}l@{}}at last, a movie that handles the probability of alien visits with \\the \textcolor{red}{appr0priate} depth and loving \textcolor{red}{warDmth}  \end{tabular} 
&  Negative \\

\begin{tabular}[c]{@{}l@{}}Word-level\\\cite{alzantot-etal-2018-generating} \end{tabular} & \begin{tabular}[c]{@{}l@{}} at last, a movie that handles the probability of alien \textcolor{red}{trips} with \\ the \textcolor{red}{adequate} depth and loving warmth \end{tabular}  
&  Negative \\

\bottomrule
\end{tabular}
\caption{\small {Examples of textual adversarial instances on IMDB and the prediction of BERT\textsubscript{BASE}} on them}
\label{tab:text_adv}
\vspace{-1em}
\end{table*}

\section{Related Work}
\label{sec:RelatedWork}

\mypar{Adversarial Defences for Image}
An intuitive adversarial defence is to train a deep neural network to be robust against adversarial input samples by e.g. mixing adversarial samples with the training data~\citep{goodfellow2014explaining, madry2017towards, xie2019feature}; popular platforms like Cleverhans \citep{papernot2016technical} are available to support robust training.
However, such defences, termed as `proactive', are expensive and vulnerable to optimisation attacks \citep{CVPRPaper}.

In contrast, others have proposed `reactive' defences that identify the variations in the representations learned by the DNN on the original input images to separate the adversarial samples; typically, these posit that adversarial examples can be characterised as belonging to particular subspaces \citep{tramer2017space}, and the different approaches aim to capture the nature of these subspaces in different ways, with detectors such as logistic regression classifiers built over the learned representations.   
\citet{feinman2017detecting} built detectors using kernel density estimation on the last hidden layer of a DNN. 
\citet{ma2018characterizing} characterised the dimensional properties of adversarial subspaces using Local Intrinsic Dimensionality (\LID), applied to the distribution of distances to neighbors in the region around a sample. 
\citet{papernot2018deep}, noting that DNNs are poorly calibrated \citep{guo-etal:2017:ICML}, proposed Deep k-Nearest Neighbors (\dknn), a \knn classifier constructed over the hidden layers of a DNN classifier; such a \dknn classifier could match the performance of the DNN while also providing better confidence estimates of prediction, and these confidence estimates are used in identifying adversarial examples.
\citet{lee2018simple} constructed Mahalanobis distance-based confidence scores from DNNs, using these scores to construct a detection classifier. 
\citet{CVPRPaper} investigated the use of influence functions in adversarial image detection that explain the decisions of a model by identifying influential training examples, and comparing these points to those found in a \dknn approach, using the differences in distributions between real examples and adversarial ones to construct classifiers that outperformed the approaches above. In this paper, we focus on the last two and adapt them to NLP.

\mypar{Adversarial Defences for Text} 
Improving adversarial robustness remains a widely used mechanism in defending textual adversaries \citep{li-etal:2016:ACL,li-etal-2017-robust,ribeiro-etal-2018-semantically,jones-etal-2020-robust}.  In NLP, however, there have been fewer reactive methods. To prevent character-level and word-level adversarial perturbations \citet{zhou-etal-2019-learning} proposed the learning to discriminate perturbations (DISP) framework that detects and replaces suspicious words. \citet{mozes-etal-2021-frequency} emphasised word frequencies in the texts in determining adversarial perturbations, arguing that adversarially infused words are less likely to occur, and constructed a rule-based, model-agnostic frequency-guided word substitutions (FGWS) algorithm. The approach of \citet{wang2022detecting} voted the prediction label for a set of samples generated by random word substitutions from a sentence and matched the voted prediction label with the original sentence’s prediction label to detect word-level adversaries. Anomaly Detection with Frequency-Aware Randomization (ADFAR) as proposed by ~\citet{bao2021defending} adds anomaly detection as an additional optimization training objective and augments the training set with random rare-frequency word substitutions of the original sentences.
Rather than focus on word substitution as the above methods, \citet{mosca2022detecting} trained an adversarial detector on Shapley additive explanations~\cite{fidel2020explainability}.

In NLP, only \citet{liu-etal-2022-detecting} has used the idea of constructing detectors over learned representations as in the image domain, which explored the idea of adapting the \LID~\citep{ma2018characterizing} method above. In addition, they proposed the MultiDistance Representation Ensemble Method (\MDRE) algorithm that puts together learned representations from multiple DNN models to detect adversarial texts. Unlike other approaches, the same detector could apply to different types of attacks (character-based, word-based, syntax-based) and MDRE in particular improved over baseline methods across the range of attacks. This motivates our adaptation of more recent techniques from the image domain.

\mypar{Influence Functions}
The influence function (IF) is a statistical method that captures the dependence of an estimator on any one of the sample (training) points. \citet{koh2017understanding} were the first to adapt IFs to image DNNs as a method for interpreting the model’s decision: the IF finds the most influential training samples, both helpful and harmful, contributing to each prediction. The essence of the approach is to consider a point $z$ from the training set and compute the change to parameters $\theta$ if $z$ were upweighted by a small $\epsilon$; they then defined closed-form expressions $\mathcal{I}(z, z_\textsubscript{test})$ to identify the most influential points $z$ on a test point $z_\textsubscript{test}$.

IFs were first applied to NLP deep architectures by \citet{han-etal-2020-explaining}, and compared with established gradient-based saliency maps as a way of interpreting input feature importance, using sentiment classification and natural language inference (NLI) as testbeds.  Their first finding was that IFs are reliable for deep NLP architectures. Their second interesting finding was that while IFs and saliency measures were consistent for sentiment classification, they differed for NLI: they concluded that for more complex understanding tasks like NLI, IFs captured more useful interpretive information.  They also found IFs to be useful for identifying and quantifying the effect of data artifacts on model prediction.
A few other works have continued investigating the usefulness of IFs in NLP, such as \citet{guo-etal-2021-fastif}, who proposed a faster method for IF computation by restricting candidates to top-$k$ nearest neighbors.

\vspace{-0.1cm}

\section{Methods}
\label{sec:method}

\vspace{-0.1cm}

\subsection{NNIF Detector}
\label{sec:nnif_method}

\vspace{-0.1cm}

We follow \citet{CVPRPaper}'s Nearest Neighbor Influence Function (NNIF) method and apply it to NLP architectures.  The essence of it is, for some point $z$ that may be regular or adversarial, to identify the training points that are most influential and those that are nearest neighbors to $z$, and to build a classifier based on those that will predict whether $z$ is regular or adversarial based on differences in relative distributions (Fig~\ref{fig:cohen_embed_space}).

We take a DNN classifier and dataset for some particular task (e.g. sentiment classification); we refer to this DNN as the \myterm{target model}.
For each test sample $z_\textsubscript{test}$, we compute the influence scores $\mathcal{I}(z, z_\textsubscript{test})$ for all training points $z$, given the target model, and select the top $M$ most helpful and $M$ most harmful (details App~\ref{app:influence_calc}). 
We then construct a \dknn classifier in the style of \citet{papernot2018deep}, using the hidden layers of the target model and 
the training points.
For each $\zt$ we find the ranks $\ra$ and distances $\di$ using this \dknn for the training examples identified by the IFs; we denote by $\ra^{M\uparrow}, \di^{M\uparrow}, \ra^{M\downarrow}, \di^{M\downarrow}$ the ranks and distances of the $2M$ most helpful and harmful training examples, respectively.
We finally construct a logistic regression classifier with features $(\ra^{M\uparrow}, \di^{M\uparrow}, \ra^{M\downarrow}, \di^{M\downarrow})$ to detect whether an input is adversarial or not. 

Where the target model of \citet{CVPRPaper} is a ResNet model, ours is a large language model (LLM) base with additional layers that are fine-tuned for the chosen tasks (\S\ref{sec:exper-target}).  The hidden layers we use for NNIF are then the pre-final additional layers on top of the DNN (\S\ref{sec:exper-methods}).

\subsection{\Mahalanobis Detector}
\label{sec:mahalanobis}

Here we follow \citet{lee2018simple}, who build a detector that captures the variation in the probability density of the class-conditional Gaussian distribution of the learned representation by the model. 
Motivated, like \citet{papernot2018deep}, by the problem that DNNs are poorly calibrated \citep{guo-etal:2017:ICML}, they replace the final softmax layer with a Gaussian Discriminant Analysis (GDA) softmax classifier.

For a set of training points $\{(x_1, y_1), . . . , (x_n, y_n)\}$ with the label $y  \in \{ 1,2,\ldots, C\}$, the class mean $\hat \mu_{c}$ and covariance $\hat {\scriptstyle\sum}$ are computed for each class $c$ to approximate the generative classifier’s parameters from the pre-trained target DNN $f(x)$. 
Next, from the obtained class-conditional Gaussian distribution, the Mahalanobis distance between a test sample $x$ and its closest distribution is measured to find the confidence score $M(x) = \max_c - (f(x) - \hat \mu_c)^T \widehat {\scriptstyle\sum}^{-1} (f(x) - \hat \mu_c)$.
Finally, we label the Mahalanobis scores for the test samples as positive and adversarial samples as negative and input this feature set to an LR detector.   

\citet{lee2018simple} propose two calibration techniques to improve the detection accuracy and make regular and out-of-distribution samples more separable: (1) \textit{input pre-processing}, where they add a small noise in a controllable manner to the test samples; and (2) \textit{feature ensemble}, which combines the confidence scores from all the hidden layers of the DNN including the final features.  Both together substantially improve the performance of the base approach; each individually reaches almost the combination of the two.
As for our NNIF detector in \S\ref{sec:nnif_method}, our target DNN will have several hidden layers, and we explore models both with final layer-only representations and feature ensembles over all hidden layers.
The input preprocessing of (1) is appropriate to the continuous space of images, but not in an obvious way to text, so we do not use that.

\vspace{-0.1cm}

\section{Experimental Setup}
\label{sec:exper}

\vspace{-0.05cm}

We broadly follow the setup of \citet{liu-etal-2022-detecting}, as the prior NLP work that has used learned representations to detect adversarial examples.

\vspace{-0.1cm}

\subsection{Tasks and Datasets}

\vspace{-0.05cm}

We work on the sentiment analysis and the natural language inference tasks, two widely tasks used in the adversarial example generation \citep{pruthi-etal-2019-combating, alzantot-etal-2018-generating, ribeiro-etal-2018-semantically,ren-etal-2019-generating,iyyer-etal-2018-adversarial, yoo-qi-2021-towards-improving, li-etal-2020-bert-attack, li-etal-2021-contextualized, jin2020bert}.  In addition, these are the two tasks that were used for the investigation of the use of influence functions in NLP \citep{han-etal-2020-explaining}.

\textbf{Sentiment Analysis} For the sentiment analysis, we use the IMDB dataset \citep{maas-etal-2011-learning} that has 50,000 movie reviews, split into 25,000 training and 25,000 test examples with binary labels indicating positive or negative sentiment. IMDB dataset has 262 words per review on average. In all experiments, we use 512 maximum sequence lengths for the language models on IMDB.

\textbf{Natural Language Inference} The Multi-Genre NLI (\Mnli) dataset~\citep{williams-etal-2018-broad}, used for the natural language inference (NLI) task, contains pairs of sentences annotated with textual entailment information. The test examples are mismatched with train examples and are collected from different sources. The dataset has 392,702 training and 9,832 testing examples labelled as three classes: entailment, neutral, and contradiction. Each text of the dataset has 34 words on average. On this dataset, we set the maximum sequence length to 256.

\subsection{Attack Methods}

We use the implementations from \citet{liu-etal-2022-detecting} of
two widely used attack methods that apply character-level and word-level perturbations to construct adversarial examples. We take a BERT\textsubscript{BASE} model (\S\ref{sec:exper-target}) as the target model. An adversarial attack is successful when the adversaries have different predictions than the target mode’s original predictions.  Our two methods are (more details in \S\ref{sec:app-attack}):

\begin{itemize}
    \item \Pruthi \citep{pruthi-etal-2019-combating}.  This is a character-level attack that tweaks the original texts by randomly swapping, dropping and adding characters or adding a keyboard mistake.

\item \Alzantot \citep{alzantot-etal-2018-generating}.  This is a word-level attack that allows the attacker to alter practically every word from the sentence if required with the context-preserving synonymous words. This implementation follows \citet{jia-etal-2019-certified} in speeding up the synonym search.

\end{itemize}

\subsection{Target Model}
\label{sec:exper-target}

Following \cite{liu-etal-2022-detecting}, we use a pre-trained BERT-base-cased model, adding a fully connected dense layer of 768 nodes, a layer of 50\% dropout, and another dense layer of 768 nodes.
The dataset split is 80-20 train-test. We train the model for 3 epochs with $5e^{-5}$ learning rate and AdamW optimization without freezing any layer of the backbone model. This BERT\textsubscript{BASE} model achieves $92.90\%$ and $82.01\%$ test accuracies on the \IMDB and \Mnli datasets respectively. The accuracies of the clean model and the model under attack are given in Table~\ref{tab:acc_clean_adv}; we note that
in all the cases, \Pruthi degrades the classifier’s performance comparatively more than \Alzantot.
Sizes for \IMDB and \Mnli datasets and number of generated adversarial texts from them are in Table \ref{tab:num_examples}.

\subsection{Detectors}
\label{sec:exper-detectors}

For data to train the adversarial example detectors on, we follow standard practice in image processing \citep{ma2018characterizing,CVPRPaper} and \citet{liu-etal-2022-detecting} and use only those examples that are correctly classified by the target model (\S\ref{sec:exper-target}) from the overall test set.  Adversarial attacks are then applied to these examples; the originals (labelled positive) and their adversarial alternatives (negative) then form the \textsc{detection dataset}.
Due to the computational intensity of estimating the influential training records for the \NNIF method, we limit our detectors to having 10k records (5k tests and 5k adversarial texts) and follow a similar data size for all the other detection methods for comparability.  
We split the detection dataset 80-20 train-test, and construct and evaluate logistic regression classifiers as detectors over this detection dataset split for our proposed methods (\S\ref{sec:exper-methods}) and baselines (\S\ref{sec:detect_methods}).

\subsection{\NNIF and Mahalanobis Methods}
\label{sec:exper-methods}

\noindent
\textbf{\NNIF}
We adapt the standard NNIF implementation of \citet{CVPRPaper}. For influence score calculation, \citet{CVPRPaper} uses the Darkon 
module for the image; we instead incorporate the influence function calculation from \citet{han-etal-2020-explaining}%
\footnote{\url{https://github.com/xhan77/influence-function-analysis}} which uses Linear time Stochastic Second-Order Algorithm \citep{agarwal2017lissa} for faster convergence, and makes several adaptations to NLP. We build the DkNN containing one layer with $l_2$ distance and brute-force search.

Because IF calculations are expensive, like \citet{CVPRPaper} and \citet{han-etal-2020-explaining} we only sample from among all neighbors: we compute the IF on 6K training datapoints uniformly randomly sampled (\citet{CVPRPaper} sample 10K neighbors from 49K training points).  We choose $M = 500$ for our main results, which is at the top end of the range of values of $M$ selected by \citet{CVPRPaper}; we show in \S\ref{sec:analysis} that, unlike the image processing domain, results in our experiments are broadly monotonically increasing as $M$ increases.

Note that we don't use the faster variant of IF computation of \citet{guo-etal-2021-fastif}, as \NNIF requires \textit{separate} perspectives from IFs and kNNs, and \FastIF restricts IF search to subsets of kNNs. 

\noindent
\textbf{\Mahalanobis}
As per \S\ref{sec:mahalanobis}, we compute the mean and covariance for each class and calculate the Mahalanobis distance score for each normal instance and its adversarial counterpart.  Like \citet{ma2018characterizing}, we consider both using only the final layer of the model and stacking scores from each layer of the model (feature ensembling).  Feature ensembling is always better, so we only include those in the main results, but do separately analyse the contribution of the feature ensembling.

\mypar{Code}
For both of these, our code uses the implementation of \citet{CVPRPaper} as a starting point
{and adapts as above.\footnote{Code: \url{https://github.com/SJabin/NNIF}.}}

\subsection{Baseline Detection Methods}
\label{sec:detect_methods}

We evaluate six adversarial text detection methods as our baseline detectors. The first four are from \citet{liu-etal-2022-detecting} (we omit the language model, as it operates essentially at the chance),
while the other two are also recent high-performing systems.\footnote{We do not include ADFAR \cite{bao2021defending}, as it works and performs similarly to (and was proposed concurrently with) RSV, but has a more complex code implementation.}
We give more details on the methods in \S\ref{sec:app-baseline}.

\noindent
\textbf{DISP} \citep{zhou-etal-2019-learning}.
This is a system that aims to correct any adversarial perturbations before an example is passed to a classifier.  \citet{liu-etal-2022-detecting} adapt this to detecting the adversarial examples.

\noindent
\textbf{FGWS} \citep{mozes-etal-2021-frequency}.
This algorithm uses a word frequency threshold and calibrated replacement approach to detect adversarial examples.  It is only designed to work against word-level attacks.

\noindent
\textbf{\LID} \citep{liu-etal-2022-detecting}.
From among image processing detection methods, \citet{liu-etal-2022-detecting} adapted the Local Intrinsic Dimensionality (LID) approach of \citet{ma2018characterizing}.  This technique creates a distribution over local distances for a test record concerning its neighbors from the training set; it then applies these to the outputs of each layer from the target model to create a detection classifier.

\noindent
\textbf{\MDRE} \citep{liu-etal-2022-detecting}.
This has similarities to \LID above but uses Euclidean distance rather than the LID measure, and creates an ensemble using different Transformer models (like \citet{liu-etal-2022-detecting}, we use BERT\textsubscript{BASE}, RoBERTa\textsubscript{BASE}, XLNet\textsubscript{BASE}, BART\textsubscript{BASE}).

\noindent
\textbf{\RSV} \citep{wang2022detecting}.
In this Randomized Substitution and Vote approach, the assumption is that a word-level attacker aims to find an optimal synonym substitution that mutually influences other words in the sentence. Hence, \citet{wang2022detecting} randomly replaces words from the text with synonyms in order to destroy the mutual interaction between words and eliminate adversarial perturbation.  Like FGWS, this is only designed to work against word-level attacks. 

\noindent
\textbf{SHAP} \cite{mosca2022detecting}.
In this approach, an adversarial detector is trained using the SHapley Additive exPlanations (SHAP) values of the training data for each test data item using the SHAP explainer \cite{fidel2020explainability}. 
They experiment on multiple classifiers as the detectors: logistic regression, random forest, support vector and neural network. In our main results, we report the best classifier for each dataset and attack.


\begin{table}
    \centering
    \small
    \captionsetup{singlelinecheck = false, justification=justified}
    \setlength\tabcolsep{1.5pt}
    \begin{tabular}{clcc}
    \toprule
    Dataset & Detector 
    & \begin{tabular}[c]{@{}l@{}}\textsc{Char}\\ \textsc{Attack}\end{tabular} 
    & \begin{tabular}[c]{@{}l@{}}\textsc{Word}\\ \textsc{Attack}\end{tabular}\\ 
    \midrule
    \multirow{9}{*}{IMDB}
    
    & \DISP* & 0.8936 & 0.7714 \\ 
    & \FGWS & --- & 0.7546 \\ 
    & \LID & 0.814 & 0.675\\ 
    & \MDRE & 0.846  & 0.7025 \\ 
    & \RSV & --- & \underline{0.8876} \\
    & \SHAP & 0.812 & 0.764 \\ 
    & \NNIF & \textbf{1.0} & \textbf{0.899} \\
    & \Mahalanobis & \underline{0.9167} & 0.8147 \\

    \midrule
    \multirow{9}{*}{\Mnli}
    
    & \DISP* & \textbf{0.7496} & 0.6137 \\ 
    & \FGWS & --- & 0.6112\\ 
    & \LID &  0.7035 & 0.5838 \\ 
    & \MDRE & 0.687 & 0.6231 \\ 
    & \RSV & --- & 0.6054 \\ 
    & \SHAP & 0.614 & \underline{0.697} \\ 
    & \NNIF & \underline{0.745} & \textbf{0.7351} \\
    & \Mahalanobis & 0.6972 & 0.6211 \\ 
    \bottomrule
    \end{tabular}
    \caption{Accuracy of detection classifiers (\textbf{best}, \underline{second}). DISP results reported from \citet{liu-etal-2022-detecting}.\vspace{-0.1cm}}
    \label{tab:result}
    \vspace{-10pt}
\end{table}

\vspace{-0.1cm}

\section{Evaluation}
\label{sec:eval}

\vspace{-0.1cm}

\subsection{Main Results}
\label{sec:results}

\vspace{-0.1cm}

Results on 
the detector baselines are in Table~\ref{tab:result}. (All \SHAP detector classifiers in Table~\ref{tab:shap_acc}.)
Overall, \NNIF is the best, performing with 100\% accuracy on \Pruthi for sentiment analysis (more than 8\% better than the second) and 90\% on \Alzantot (more than 1\% better than the second, RSV, which is tailored to word-level attacks).
For \Mnli \Alzantot, it is around 4\% better than the second best.  The only one where it is not best, \Pruthi, is only very slightly below the best performer DISP.  (We note that for DISP we report the accuracy values from \citet{liu-etal-2022-detecting}.  This means that the DISP detector used more data in its training set, and so has an advantage in this respect.)   
\Mahalanobis also performs quite strongly, either better or similar to the baseline detectors, although not as strongly as \NNIF; this mirrors the findings in image processing.  \MDRE results are lower than in \citet{liu-etal-2022-detecting} as a consequence of using less data for training all detection classifiers, as discussed in \S\ref{sec:exper-detectors}.

In terms of aggregate task performance, in all our experiments, the detection accuracy on the natural language inference task is lower than the sentiment analysis task in general. As the \Mnli dataset is a three-class problem and additionally uses mismatched test sentences, the detection is innately harder.

\begin{figure}
    \centering
    \captionsetup{singlelinecheck = false, justification=justified}
    \includegraphics[width=.45\textwidth]{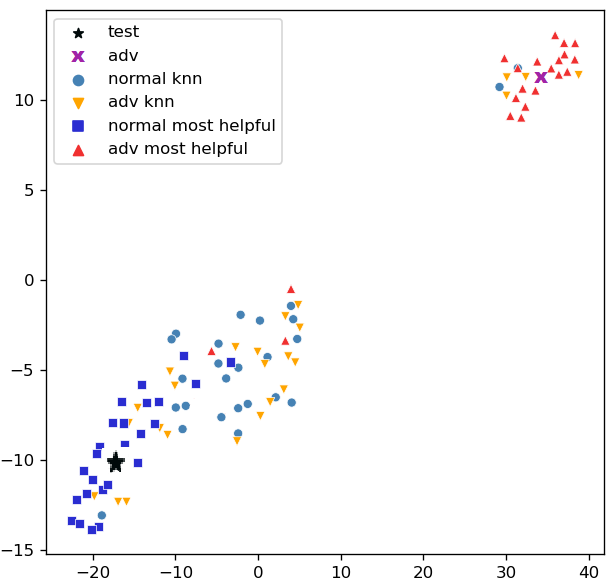}
    \caption{The correspondence between the helpful training records based
on IFs in the embedding space of a DNN trained on the \IMDB dataset. We present (using \tsne) the embedding space of a DNN for an actual example (black star) with its adversarial version (purple cross) along with their 25 nearest neighbors (blue) and most helpful samples based on the IF (red).\vspace{-0.1cm}}
    \label{fig:subspace_nlp}
\vspace{-1em}
\end{figure}

\begin{table}
\centering
\small
\captionsetup{singlelinecheck = false, justification=justified}
\resizebox{\columnwidth}{!}{%
\begin{tabular}{llcc}
\toprule
Dataset & Attack & \begin{tabular}[c]{@{}c@{}}Penultimate \\ layer\end{tabular} & \begin{tabular}[c]{@{}c@{}}Feature\\ Ensemble\end{tabular} \\ \midrule
\multirow{2}{*}{\IMDB} & \Pruthi   & 0.5967 & 0.9167 \\
                      & \Alzantot & 0.536  & 0.8147 \\\midrule
\multirow{2}{*}{\Mnli} & \Pruthi   & 0.5172 & 0.6972 \\
                      & \Alzantot & 0.4983 & 0.6212 \\\bottomrule
\end{tabular}%
}
\caption{Detection accuracy of Mahalanobis detector in two settings: penultimate layer (no calibration) and feature ensemble.\vspace{-0.1cm}}
\label{tab:mahalanobis_acc}
\end{table}

\vspace{-1em}

\subsection{Analyses}
\label{sec:analysis}

\vspace{-0.1cm}

\mypar{Regions around adversarial examples}
The assumption underpinning the \citet{CVPRPaper} method is that influential training samples and nearest neighbors should overlap for normal examples, but less so for adversarial examples: having two views on `nearby' points is key, illustrated in Fig~\ref{fig:cohen_embed_space}. We produce an analogous figure in Fig~\ref{fig:subspace_nlp} for a randomly selected IMDB test point and its adversarial counterpart generated by \Alzantot. We plot 25 nearest neighbors and 25 most helpful IF points using \tsne \citep{JMLR:v9:vandermaaten08a}.  
Ideally, normal neighbors and influence points (blue) should be more tightly grouped and closer to the test point (star); \citet{CVPRPaper} expect that for the adversarial point (cross), the neighbors (orange down triangle) should often be separated from the influence points (red up triangle). We see this to some extent in Fig~\ref{fig:subspace_nlp} with many adversarial neighbors near the normal point but adversarial influence points near the adversarial point.

\begin{figure}
    \centering
    \captionsetup{singlelinecheck = false, justification=justified}
    \includegraphics[width=.4\textwidth]{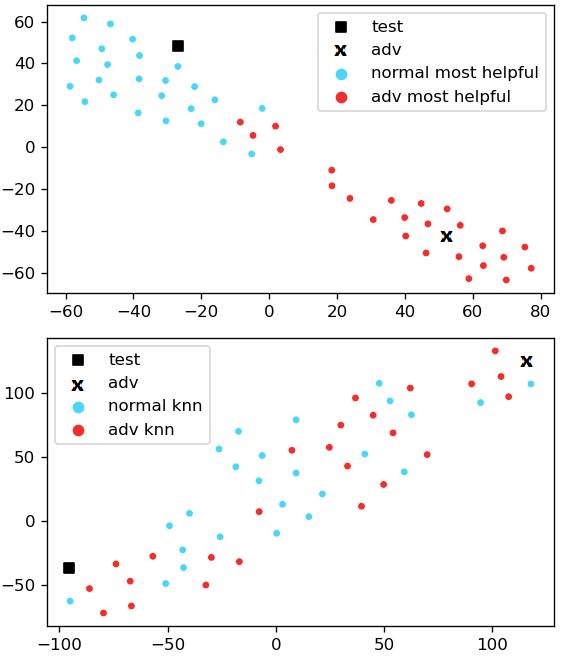}
    \caption{Normal and adversarial train subspace observed on the \IMDB record used in Fig \ref{fig:subspace_nlp} under \Alzantot by influence function (top) and \dknn (bottom)}
    \label{fig:imdb_syn_tsne}
    \vspace{-1em}
\end{figure}

This is more difficult to see than in the idealised schematic of Fig~\ref{fig:cohen_embed_space}, so for one view of differences in this pair of points we separate IFs and NNs in Fig~\ref{fig:imdb_syn_tsne} with recalculated \tsne for each. It is apparent that the IFs by themselves do a good job of separating normal from adversarial examples here, while the NNs are more mixed.  We give representative examples for the other datasets and attacks in App~\ref{app-subspace}.  The same pattern is true for the IMDB example on \Alzantot.  For both \Mnli, however, the IFs are less clearly separating the points, so the NNIF method relies on combining the two (NN, IF) views in the detector.

To verify whether this is more generally true than just visually for Fig~\ref{fig:imdb_syn_tsne}, we aim to measure how separable the samples of these plots are.  
As a measure of separability, we train 2000 SVC binary classifiers, one for each of our 1000 sampled test and adversarial point pairs, for both IFs and NNs.  Each classifier is trained using GridsearchCV on the top 100 points in \tsne space (either IFs or NNs), so each classifier corresponds to a plot like those in Fig~\ref{fig:imdb_syn_tsne} (App~\ref{sec:app-sep}). Accuracies averaged across the 1000 classifiers are in Table~\ref{tab:svm_acc}, with $p$-values for a one-tailed test of proportions (positing alternative hypothesis $H_1$ 
that the IF classifier is more accurate).  Table~\ref{tab:svm_acc} indicates that the IF points are generally much more clearly separable and so IF points contribute especially strongly to the method, except for \Mnli against \Alzantot, where they are essentially the same and the method relies on the two-view aspect of \NNIF.  
This observation about the relative importance of the IF contribution was not made by \citet{CVPRPaper}, and so may be specific to NLP tasks, although this would require more investigation to verify.
We also note that our results align with observations of \citet{han-etal-2020-explaining}, that in the harder task of \Mnli (\S\ref{sec:results}, Table~\ref{tab:svm_acc}), IFs provide a different perspective to characterising the datapoint of interest. We give some text examples in App~\ref{sec:app-results}.

\begin{figure}
    \centering
    \captionsetup{singlelinecheck = false, justification=justified}
    \includegraphics[width=0.4\textwidth]{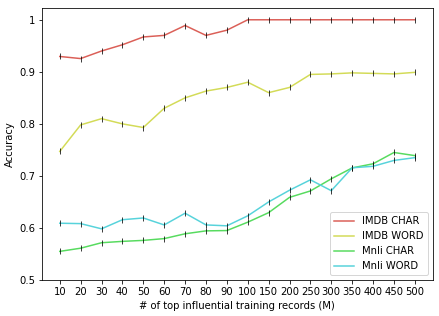}
    \caption{Accuracy of \NNIF for different values of M.}
    \label{fig:nnif_vs_m}
    \vspace{-1em}
\end{figure}

\begin{table}
\centering
\small
\setlength\tabcolsep{1pt}
\captionsetup{justification=justified}
\begin{adjustbox}{max width=.49\textwidth}
\begin{tabular}{@{}lccr@{}}
\toprule
    Attack &
    \begin{tabular}[c]{@{}c@{}}Avg Acc\\ NNIF\end{tabular} &
    \begin{tabular}[c]{@{}c@{}}Avg Acc\\ \knn\end{tabular} &
 $p$-value \\
\midrule
    \IMDB$_\Pruthi$   & 0.6875 & 0.5626 & $< .00001$\\
    \IMDB$_\Alzantot$ & 0.7812 & 0.5644 & $< .00001$\\
\midrule
    $\Mnli_\Pruthi$   & 0.6399 & 0.5625 & $< .00001$\\
    $\Mnli_\Alzantot$ & 0.5603 & 0.5632 & 0.448\\
\bottomrule      
\end{tabular}
\end{adjustbox}
\caption{SVC accuracy of linearly separating the 2D-\tsne embedding subspace of neighboring train samples of 1000 test records and their adversarial versions}

\label{tab:svm_acc}
\vspace{-1em}
\end{table}

\begin{figure}[t]
    \centering
    \captionsetup{singlelinecheck = false, justification=justified}
    \includegraphics[width=.4\textwidth]{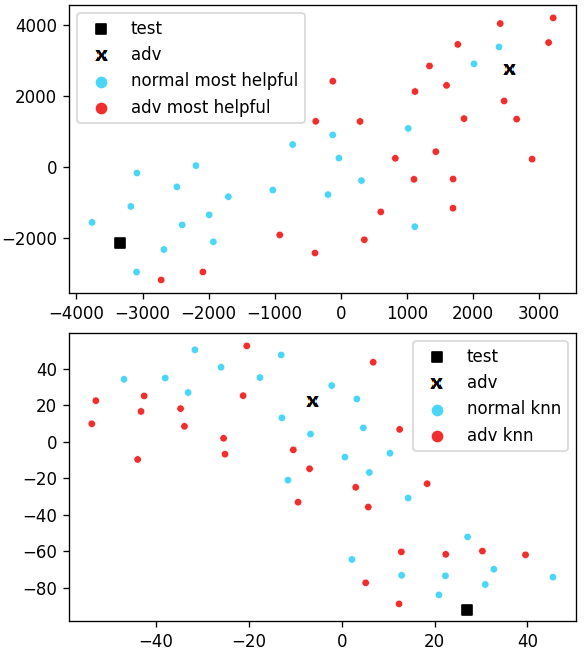}
    \caption{Normal and adversarial subspace of the \Mnli \Pruthi text in Table \ref{tab:example_nnif_word} by IF (top) and \dknn (bottom)}
    \label{fig:mnli_tab11}
    \vspace{-1em}
\end{figure}

\begin{figure}[t]
    \centering
    \captionsetup{singlelinecheck = false, justification=justified}
    \includegraphics[width=.4\textwidth]{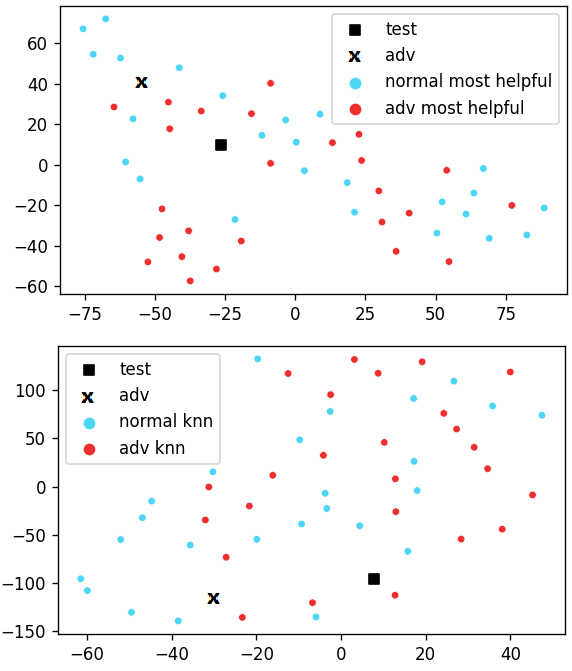}
    \caption{Normal and adversarial subspace by IF (top) and \dknn (bottom) on the unsuccessful detection by \NNIF of the \Mnli \Pruthi text in Table \ref{tab:example_nnif_incorr_char}}
    \label{fig:mnli_tab12}
    \vspace{-1em}
\end{figure}

To look further into the more challenging combination of \Mnli and \Pruthi (as the one case in Table~\ref{tab:result} where NNIF was not the highest scoring, albeit by a small margin), we consider a successful and an unsuccessful detection case by \NNIF, with the actual examples given in the appendices in Tables~\ref{tab:example_nnif_char} and \ref{tab:example_nnif_incorr_char}, and the corresponding \tsne plots of IF and NNs in Figs~\ref{fig:mnli_tab11} and \ref{fig:mnli_tab12}, respectively.  The IFs in Fig~\ref{fig:mnli_tab11} (the correct example) are somewhat more clustered, with the red (adversarial) points mostly in the top right, than the IFs in Fig~\ref{fig:mnli_tab12} (the incorrect example); this lines up with the results of Table~\ref{tab:svm_acc} in that separability of IF does seem to matter for \Mnli+\Pruthi.

\mypar{Varying $M$ in \NNIF}
Fig~\ref{fig:nnif_vs_m} plots the accuracies of the \NNIF method for both tasks and attacks, for a range of values of $M$.  The accuracy broadly monotonically increases until plateaus for the IMDB results, although the \Mnli results look to be still increasing.  This is a contrast with the image processing results of \citet{CVPRPaper}, where much smaller values of $M$ (e.g. 30) produced better results.  It is unclear what characteristics of our tasks (fewer classes, more long-distance dependencies, \ldots) lead to this difference.

\mypar{Ablation for \Mahalanobis}
Table~\ref{tab:mahalanobis_acc} shows the accuracies of \Mahalanobis using only the final layer or the feature ensemble. As with \citet{lee2018simple}, the feature ensemble produces much better results. The improvement is larger for IMDB, but still important for \Mnli, as without the ensemble, detection is essentially at the chance. Noting that the target model of \citet{lee2018simple} had many more hidden layers in the ensemble, it is an open question as to whether introducing additional dense layers into our LLM-based model might improve detection while still preserving target model performance.

\section{Conclusion and Future Work}

We have adapted from image processing two methods,  \NNIF \citep{CVPRPaper} and \Mahalanobis \cite{lee2018simple}, that detect adversarial examples using learned representations.  Both perform strongly, with \NNIF the best on three of four task/attack combinations, and a close second on the fourth, against several strong baselines.

Our analysis shows that influence function points make a particularly important contribution to the \NNIF method.  The \Mnli task is more challenging for all methods; here it is the complementary nature of information from influence functions and nearest neighbors, supporting observations by \citet{han-etal-2020-explaining} about the different perspective of influence functions in this more complex NLP task.

The \NNIF method is computationally expensive, so future work will look at ways to make it more efficient.  Additionally, to gain a fuller understanding of what information influence functions can provide in NLP tasks, future work will look at a wider range of tasks and attacks.

\section{Limitations}

The major limitation is the computationally expensive calculation of influence functions in our \NNIF method. For this, following \citet{CVPRPaper} we restrict the data size to 10k (5k test, 5k adversarial) for \NNIF and follow a similar approach for other methods for comparability. This helps faster explanation generation in \SHAP as well. We use a small architecture as recommended in \citet{han-etal-2020-explaining} for the BERT\textsubscript{BASE} model for \NNIF and other detectors. 
As noted in the paper, we recognise that there is the \FastIF method of \citet{guo-etal-2021-fastif} for speeding up influence function calculation, but because of the restriction of influence function points to nearest neighbors, it is not suitable for our application.

We use only two datasets/tasks and two attack methods, partly because of the computational expense of \NNIF. While they are commonly used in the adversarial example literature as well as the analysis of influence functions in NLP by \citet{han-etal-2020-explaining} and represent different levels of task complexity and attack type, a wider range of datasets/tasks and attack methods is needed for a full characterisation of influence functions and the nature of adversarial subspaces.

For all experiments, we restrict the maximum sequence length following \citet{liu-etal-2022-detecting}, which may influence the detectors' performance, especially for the NLI task, that requires the model to learn from a hypothesis and premise text pairs.

For the detector baselines, we used the most available methods. There are two recent contemporaneous methods by \citet{wang2022detecting} and \cite{bao2021defending} that explore the idea that adversarial perturbations are typically rare-frequency words, and create augmented training sets by replacing those words in each sentence with synonyms. For the detection, \citet{wang2022detecting} matches the voted prediction with the obtained prediction and \cite{bao2021defending} trains the model on a separate auxiliary learning objective. Between these two works, we choose the \RSV from \citet{wang2022detecting} in our work.
For \RSV, we follow the similar setting from \citet{wang2022detecting} in choosing the vote number, word substitution rate and stop word selection for both \IMDB and \Mnli. A different setting for \Mnli may improve the result.

\setlength{\bibsep}{0pt}

\bibliography{anthology,custom_short, custom}
\bibliographystyle{acl_natbib}

\appendix
\newpage

\section{Experimental Setup Details}
\label{sec:app-exper-setup}
The size of the datasets and the number of adversarial samples generated by each of the attack methods are given in Tab. \ref{tab:num_examples}. Obtained accuracies of the BERT\textsubscript{BASE} model are in Tab. \ref{tab:acc_clean_adv} and the other models used in \MDRE are in Tab. \ref{tab:acc_mdre_lm}

\begin{table*}
    \centering
    \small
    \setlength\tabcolsep{1.5pt}
    \begin{tabular}{ccccccc}
    \toprule
    \multirow{2}{4em}{\centering \textbf{Dataset}} & \multirow{2}{4em}{\centering \textbf{Training.}} & \multirow{2}{5em}{\centering \textbf{Validation.}} &\multirow{2}{4em}{\centering \textbf{Testing.}} & \multirow{2}{9em}{\centering \textbf{Correctly Predicted Test Examples}} & \multicolumn{2}{c}{\textbf{Adversarial/Original Examples}} \\ [0.5ex]
    & & & & &  \multicolumn{1}{c}{character-level} & \multicolumn{1}{c}{word-level} \\ 
    \midrule
    IMDB & 20,000 & 5,000 & 25,000 & 23,226 & 12,299 & 9,627 \\
    \Mnli & 314,162 & 78,540 & 9,832 & 8,062 & 7,028 & 3,240 \\
    \bottomrule
    \end{tabular}
    \caption{\small{The number of examples used in experiments}}
    \label{tab:num_examples}
    \vspace{-10pt}
\end{table*}

\subsection{Attack Methods}
\label{sec:app-attack}

\par{\textbf{\Pruthi.}} We implement \Pruthi as proposed by \citet{pruthi-etal-2019-combating}. It tweaks the original texts by randomly swapping, dropping and adding characters or adding a keyboard mistake. \textit{Swapping} refers to exchanging places of two adjacent internal characters. \textit{Dropping} removes a character and \textit{Adding} inserts a new character at a randomly selected position. \textit{Keyboard mistakes} is for substituting a character with one of its adjacent characters in keyboards.

In our experiments, we allow a maximum of half the words from the original text to be perturbed, so the maximum number of possible attacks on the \IMDB and \Mnli datasets is 256 and 128 per sentence, respectively.

\par{\textbf{\Alzantot.}}
\citet{alzantot-etal-2018-generating} proposed an effective and widely used adversarial attack that we incorporate in our work as \Alzantot.

This method allows the attacker to alter practically every word from the sentence if required with the context-preserving synonymous words. The synonym search is done over a large search space that includes the GloVe word vectors \citep{pennington-etal-2014-glove}, counter-fitting word vectors \citep{mrksic-etal-2016-counter}, and the Google 1 billion words language model \citep{chelba2013one}.  Then, following the natural selection methods, crossover and mutation techniques from the population-based genetic algorithm are applied to generate the next set of adversarial sentences. On each iteration, several adversarial texts that are unsuccessful in changing the model’s prediction are removed from the pool.

However, \citet{jia-etal-2019-certified} found that the algorithm is computationally expensive and recommended using a faster language model and stopping the semantic drift of the algorithm that refers to applying the language model on the synonyms picked from previous iterations as well to choose words from their neighboring word-space. 

We incorporate the above recommendations by utilising a faster Transformer-XL architecture~\citep{dai-etal-2019-transformer} that is pretrained on the WikiText-103 dataset \citep{merity2016pointer} and prohibiting the semantic drift by finding all test examples words' neighbors only before attacks. We also restrict the minimum number of perturbations to one-fifth of the maximum sequence length which is 102 and 51 for the \IMDB and \Mnli, respectively. 

\begin{table}[ht]
\centering
\small
\setlength\tabcolsep{1.5pt}
\begin{tabular}{@{}lccc@{}}
\toprule

\multirow{2}{3em}{Dataset} & Clean & \textsc{Char} & \textsc{Word} \\
                        & Accuracy & \textsc{Attack} &\textsc{Attack} \\ [0.5ex]
\midrule
\IMDB & 0.9290  & 0.3656  & 0.6999 \\
\Mnli & 0.8201  & 0.4848  & 0.6864  \\ 
\bottomrule
\end{tabular}%
\caption{BERT\textsubscript{BASE} classifier accuracy on the clean and adversarial examples}
\label{tab:acc_clean_adv}
\end{table}

\begin{table*}
    \centering
    \small
    \setlength\tabcolsep{1.5pt}
    \captionsetup{justification=centering}
    \begin{tabular}{llcccc}
    \toprule
    Dataset & Attack Method & BERT\textsubscript{BASE} & RoBERTa\textsubscript{BASE} & XLNet\textsubscript{BASE} & BART\textsubscript{BASE} \\ [0.5ex]
    \midrule
    \multirow{3}{5em}{\IMDB} & Clean & 0.9290 & 0.9532 & 0.9336 & 0.9429 \\
    & \Pruthi & 0.3656  & 0.8613  & 0.5770  & 0.8286  \\
    & \Alzantot &  0.6999 &  0.8714 & 0.7918  &   0.8425 \\
    
    \midrule
    \multirow{3}{5em}{\Mnli} & Clean & 0.8201 & 0.8671 & 0.8630 & 0.8455 \\
    & \Pruthi & 0.4848 &  0.7104 & 0.6670  & 0.6457  \\
    & \Alzantot & 0.6864
  & 0.7068  & 0.6870  &  0.6296  \\
  
    \bottomrule
    \end{tabular}
    \caption{\small{Different classifier accuracies on both clean and adversarial dataset for \MDRE.}}
    \label{tab:acc_mdre_lm}
    \vspace{-10pt}
\end{table*}

\subsection{Baseline Detection Methods}
\label{sec:app-baseline}

The first four are from \citet{liu-etal-2022-detecting} (we omit the language model, as it operates essentially at the chance), and we use the implementations from there.%
\footnote{\url{https://github.com/NaLiuAnna/MDRE}}

\par{ \textbf{Learning to Discriminate Perturbations (DISP) \citep{zhou-etal-2019-learning}.}}
DISP is one of the commonly used baselines for adversarial text detection that identifies a set of character-level of word-level perturbed tokens and then applies an embedding estimator that predicts embeddings for each perturbed token and maps them to the actual word to repair the perturbations.

If the model’s prediction on an adversarial text restored by DISP remains the same class as the prediction on its original version, we consider it a successful detection of an adversarial example.

\par{\textbf{Frequency-guided word substitutions \\ (FGWS)~\citep{mozes-etal-2021-frequency}.}}
\citet{mozes-etal-2021-frequency} verifies that in the case of word-level attacks, the synonym replacements normally occur in low frequency. They use this concept in a model-agnostic rule-based adversarial text detection algorithm Frequency-Guided Word Substitutions (FGWS).  

Firstly, the algorithm sets a word frequency threshold to identify infrequent words that have frequencies lower than this value. Then the algorithm replaces those words with their high-frequency synonyms and selects the replaced sentences as adversarial samples if the model’s prediction confidence scores for the replacements change over a threshold. They use WordNet \citep{Fellbaum2005-FELWAW} and GloVe vectors \citep{pennington-etal-2014-glove} to find the synonyms. They experiment by taking~\{$\mathit{0}$-th, $\mathit{10}$-th, $\cdots$, $\mathit{100}$-th\} percentile of word frequencies in the training set as the word-frequency threshold. Finally, on these selected alternative sentences, if the prediction confidence differs from their corresponding original sentence’s prediction confidence by more than a certain amount, the original sentences are determined as adversarial examples.

\par{ \textbf{Local Intrinsic Dimensionality (\LID)~\citep{liu-etal-2022-detecting}.}}
From the image processing detection methods, \citet{liu-etal-2022-detecting} adapt the Local Intrinsic Dimensionality (LID) approach of \citet{ma2018characterizing}.  This technique creates a local distance distribution for a test record to its neighbors from the training set. They apply this to transformer models by taking the outputs of each layer from the target model to represent the training records.

Following~\citet{liu-etal-2022-detecting}, we use the BERT\textsubscript{BASE} model and implement a logistic regression classifier as the detector, and tune the size of the neighbors $k$ through a grid search over 100, 1000, and the range [10, 42) with a step size 2.

\par{ \textbf{MultiDistance Representation Ensemble Method (\MDRE)~\citep{liu-etal-2022-detecting}.}}
Motivated by the notion that adversarial examples are out-of-distribution samples as recognized in~\citet{lee2018simple} and \citet{feinman2017detecting}, \citet{liu-etal-2022-detecting} assume that texts with the same prediction label lie on similar data submanifold and adversarial perturbation on these texts put them to another data submanifold, thus altering the model’s prediction on them. 

They measure the Euclidean distance between each reference datapoint and the nearest neighbors from the training datapoints with similar predicted labels and establish that this distance will be greater for the adversarial reference point than the normal one. They further use ensemble learning to combine distances between representations learned from multiple DNNs and build a binary logistic regression model to detect adversarial examples.

Following~\citep{liu-etal-2022-detecting}, we also use four learning models: [BERT\textsubscript{BASE}, RoBERTa\textsubscript{BASE}, XLNet\textsubscript{BASE}, BART\textsubscript{BASE}] in our experiments.
Table~\ref{tab:acc_mdre_lm} reports the clean accuracies of the other target classifiers used in feature ensembling in \MDRE.

\par{\textbf{Randomized Substitution and Vote (\RSV)~\citep{wang2022detecting}.}}
A word-level attacker’s target is to find an optimal synonym substitution that mutually influences other words in the sentence. Taking this optimization target of the adversary, \citet{wang2022detecting} resort to randomly substituting words from the text with their synonyms and argue that this random word substitution destroys the mutual interaction between words and eliminates adversarial perturbation.

At first, they generate a set of perturbed samples by randomly replacing some words from a text with their arbitrary synonyms. Then the model’s output logits for the processed samples are accumulated and voted to determine a prediction label for the text samples. If the original text’s prediction doesn’t match the voted prediction label it is considered as an adversarial example.

We use their code.\footnote{\url{https://github.com/JHL-HUST/RSV}}

\par{\textbf{SHapley Additive
exPlanations (SHAP)~\cite{mosca2022detecting}.}}
In this work, \citet{mosca2022detecting} adopt an adversarial image detection method for word-level attacks on text. They train an adversarial detector with the SHapley Additive exPlanations (SHAP) values of the training data for each of the test data using the SHAP explainer proposed and implemented by \citet{fidel2020explainability}. 

They experiment on multiple classifiers as the detectors such as logistic regression, random-forest classifier, support vector classifier and a neural network. They also show that the detector doesn’t require a large number of training samples for it to be successful. In our work, we follow the same and report the best accuracy obtained among the four detectors. 

We use their code.\footnote{\url{https://github.com/huberl/adversarial_shap_detect_Repl4NLP}} Accuracies of all the detectors are in Table~\ref{tab:shap_acc}.

\begin{table}
\centering
\small
    \setlength\tabcolsep{1.5pt}
    \captionsetup{justification=centering}
\resizebox{\columnwidth}{!}{%
\begin{tabular}{@{}llcccc@{}}
\toprule
Dataset & Attack & \begin{tabular}[c]{@{}c@{}}Logistic\\ Regression\end{tabular} & \begin{tabular}[c]{@{}c@{}}Random \\ Forest\end{tabular} & SVC & DNN \\
\midrule
\multirow{2}{*}{\IMDB} & \Pruthi   & 0.740 & 0.804 & 0.803 & 0.812 \\
                      & \Alzantot & 0.605 & 0.764 & 0.684 & 0.75  \\
\midrule
\multirow{2}{*}{\Mnli} & \Pruthi   & 0.588 & 0.614 & 0.613 & 0.61  \\
                      & \Alzantot & 0.528 & 0.697 & 0.633 & 0.621\\
\bottomrule
\end{tabular}%
}
\caption{Detection accuracy obtained from four detector classifiers used in \SHAP.}
\label{tab:shap_acc}
\end{table}

\section{Computing Influence Function}
\label{app:influence_calc}
For a datapoint $z_i = (x_i, y_i)$ from the training set $\{(x_1, y_1),\ldots,(x_i, y_i) \in (X,Y)\}$ and model parameters $\theta \in \Theta$, the loss of the model be $L(z,\theta)$ and the optimized parameters are:
\[{\theta'} = \argmin_{\theta\in\Theta} \frac{1}{n} \sum_{i=1}^{n} L(z_i, \theta)\]

The influence score is then calculated by observing the impact of a modification in the weight of a train datapoint on the decision of the prediction for the test datapoint. Assume we upweigh the training datapoint $z$ by a small $\epsilon$ amount, which produces below $\theta'$:

\[{\theta'}_{\epsilon,z} = \argmin_{\theta\in\Theta} \frac{1}{n} \sum_{i=1}^{n} L( z_i, \theta) + \epsilon L(z, \theta)\]

Then, according to \citet{koh2017understanding}, the influence of the boosted $z$ on the parameters $\theta'$ can be defined by:

\begin{equation}
\label{eq:form1}
    \frac{d\theta'_{\epsilon,z}}{d\epsilon} |_{\epsilon=0} = - H^{-1}_{\theta'} \nabla_{\theta} L(z, \theta')
\end{equation}

where $H_\theta' = \frac{1} {n} \sum_{i=1}^{n} \nabla_{\theta}^2 L(z_i,\theta')$ is the Hessian of the model. 

Applying the chain rule to the Eq.~\ref{eq:form1} can be derived to the below form that measures the influence $I_{up,loss}$ of $z$ on the loss of a test point $z_{test}$:
    \begin{multline}
    \label{eq:influence_score}
    I_{up,loss} (z,z_{test}) = \\
    -\nabla_\theta L(z_{test},\theta') H^{-1}_{\theta} \nabla_{\theta} L(z,\theta')
    \end{multline}

The \NNIF method uses the $I_{up,loss}$ score.

\begin{table}
\centering
\begin{tabular}{@{}ll@{}}
\toprule
Parameters & Values                                   \\ \midrule
C          & {[} 1, 10, 1000, 10000{]}                \\
gamma      & {[}1, .1, .01, 0.001, 'auto'{]}          \\
kernel     & {[}'linear', 'rbf', 'poly', 'sigmoid'{]} \\ \bottomrule
\end{tabular}
\caption{Gridsearch parameters for building SVC.}
\label{tab:gridsearch}
\end{table}

\section{Illustrations of Regions Around Test and Adversarial Points}
\label{app-subspace}

\begin{figure*}
    \centering
    \begin{subfigure}[b]
    {0.45\textwidth}
    \includegraphics[width=\textwidth]{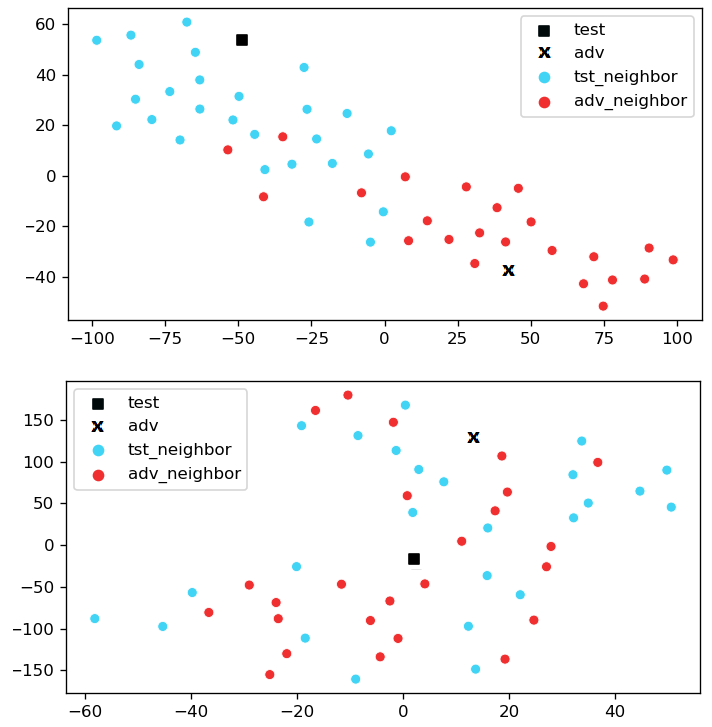}
    \caption{\Pruthi}
    \end{subfigure}
    \hspace{.2em}
    \begin{subfigure}[b]
    {0.45\textwidth}
    \includegraphics[width=\textwidth]{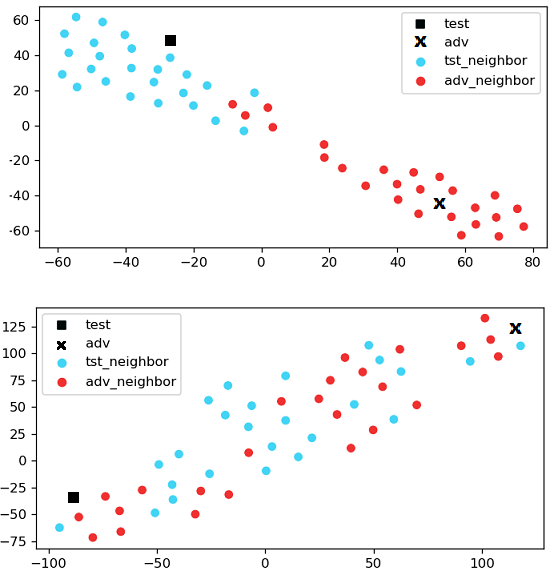}
    \caption{\Alzantot}
    \end{subfigure}
     \captionsetup{singlelinecheck = false, justification=justified}
    \caption{Embedding subspace (applied \tsne) of a test sample from the \IMDB dataset (black-square) and its adversarial version (purple-cross) generated by three types of attacks. The top set of images shows the 25 most influential training samples and the bottom set shows the top 25 nearest neighbors (\knn).}
    \label{fig:tsne_single_imdb}
    \vspace{-2cm}
\end{figure*}

\begin{figure*}
    \centering
    \begin{subfigure}[b]{0.45\textwidth}
    \includegraphics[width=\textwidth]{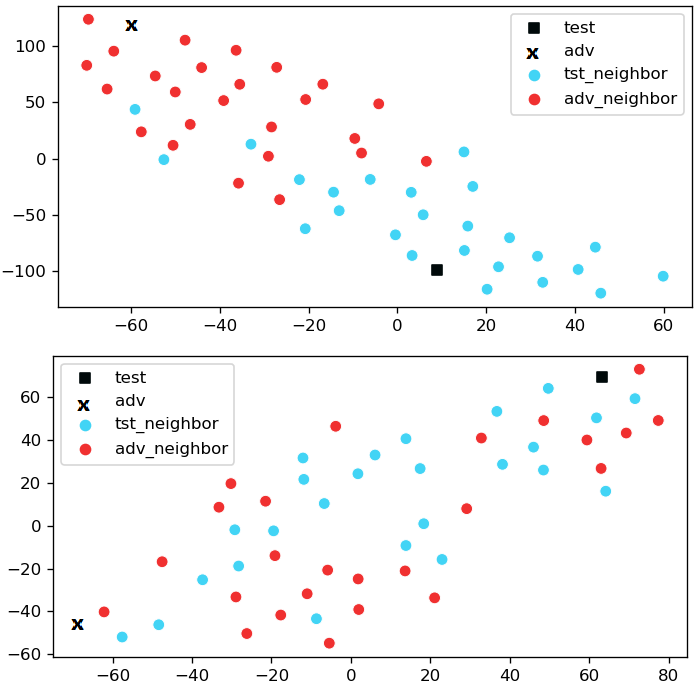}
    \caption{\Pruthi}
    \end{subfigure}
    \hspace{.2em}
    \begin{subfigure}[b]{0.45\textwidth}
    \includegraphics[width=\textwidth]{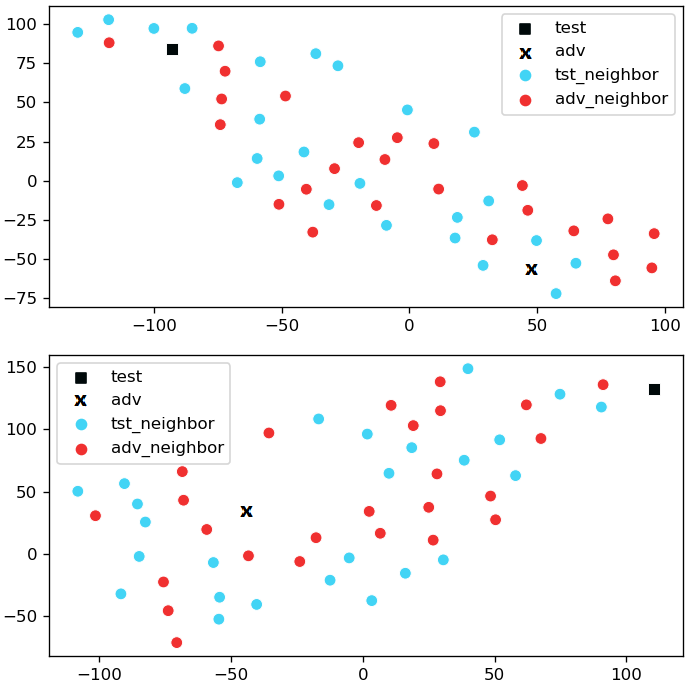}
    \caption{\Alzantot}
    \end{subfigure}

    \captionsetup{singlelinecheck = false, justification=justified}
    \caption{Embedding subspace (applied \tsne) of a test sample from the \Mnli dataset (black-square) and its adversarial version (purple-cross) generated by three types of attacks. The top set of images shows the 25 most influential training samples and the bottom set shows the top 25 nearest neighbors (\knn).}
    \label{fig:tsne_single_mnli}
    \vspace{-2cm}
\end{figure*}

Looking at the training samples that influence the prediction of a test datapoint, gives us an illustration of the decision subspace of the DNN on it. To illustrate the subspace, we measure the top 25 influential (IF) and nearest neighbor (NN) training embeddings for a test record and its adversarial counterpart for each attack and plot them along with the test and adversarial points. All embeddings are reduced to two dimensions by using \tsne. Figures~\ref{fig:tsne_single_imdb} and \ref{fig:tsne_single_mnli} show an example each for the \IMDB and \Mnli datasets, respectively. On each figure, the top row depicts the IF-based training points and the bottom row shows the NN-based training points. %

\section{Separability of Points: IF vs NN}
\label{sec:app-sep}

We build SVC classifiers on the neighboring train embeddings to evaluate how well the influence function is describing the learned subspace of the DNN than the \dknn. The best SVC classifiers over NNs and IF points for each of the 1000 test and adversarial example pairs are estimated through GridSearch over the parameters as depicted in Table~\ref{tab:gridsearch}. 

\section{Experimental Results Examples}
\label{sec:app-results}
\NNIF combines the \dknn ranking on top of the influence scores to select the best training instances for a test datapoint. In Tables~\ref{tab:example_nnif_word}, \ref{tab:example_nnif_char} and \ref{tab:example_nnif_incorr_char} we illustrate examples for \Alzantot and \Pruthi respectively, showing the top three helpful and harmful training instances for the detection of the adversarial attack. We also show the \dknn rankings of the top training instances filtered by the IF scores in the table. 

As \DISP performs better in one of the experimental settings in \citet{liu-etal-2022-detecting}, we further pick one example sentence from the paper that \DISP detects correctly and observe \NNIF's performance on it. \NNIF is also able to detect the sentence correctly. In Table~\ref{tab:example_nnif_disp} we show the influential instances for this prediction as well.

\begin{table*}[t]
\centering
\normalsize
\captionsetup{justification=justified}
\resizebox{\textwidth}{!}{
\begin{tabular}{lc}
\toprule
\multicolumn{2}{l}{\textbf{Original text - label Entailment - prediction Entailment}} \\
\midrule
\multicolumn{2}{l}{\begin{tabular}[c]{@{}l@{}}\textbf{Premise:} Address your remarks to the chair  illustrates metonymy  a figure of speech in which something  is called by the name of something \\else associated with it.\end{tabular}} \\
\multicolumn{2}{l}{\textbf{Hypothesis:} Using one word to refer to something that is associated with it is a figure of speech.} \\

\midrule
\multicolumn{2}{l}{\textbf{\Alzantot - prediction Contradiction}} \\
\midrule
\multicolumn{2}{l}{\begin{tabular}[c]{@{}l@{}}\textbf{Premise:} Address your remarks to the chair \textcolor{red}{draws} metonymy a \textcolor{red}{digit} of speech in which something is called by the name of something \\else associated with it .\end{tabular}} \\
\multicolumn{2}{l}{\textbf{Hypothesis:} Using one word to refer to something that is associated with it is a \textcolor{red}{digit} of speech} \\

\midrule
\textbf{\textcolor{blue}{Top Helpful}} & \textbf{\textcolor{blue}{NNIF Rank}}\\

\midrule

\begin{tabular}[c]{@{}cl@{}} 1 & \textbf{Premise:} my adult women friends are anguishing over over some of these choices.\\
&  \textbf{Hypothesis:} I don't have any woman friends.\end{tabular} &
  2185 \\

\midrule
\begin{tabular}[c]{@{}cl@{}} 2 & \textbf{Premise:} Vrenna, now! Jon kicked the barrel and it broke open.\\
& \textbf{Hypothesis:} Jon told Vrenna what to do. \end{tabular} &
  4956 \\

\midrule
\begin{tabular}[c]{@{}cl@{}}3 & \textbf{Premise:} I brought my Gauntlet to bear; electricity leaping out. \\
& \textbf{Hypothesis:} My gauntlet was magical and electricity jumped out of it.
\end{tabular} &
  5473 \\
  
\midrule
\textbf{\textcolor{orange}{Top Harmful}} & \textbf{\textcolor{orange}{NNIF Rank}}\\

\midrule
\begin{tabular}[c]{@{}cl@{}} 1 & \textbf{Premise:} Examining the elements of the definition also may help make this distinction clear.\\
& \textbf{Hypothesis:} Ignoring the elements of the definition also may help make this distinction clear.
\end{tabular} &
 4308  \\

 \midrule
\begin{tabular}[c]{@{}cl@{}} 2 & \textbf{Premise:}
the uniformed services, recognize that promotional material received by a uniformed service member traveling on official \\ & business at government expense belongs to the government and must be relinquished in accordance with service regulations.\\
& \textbf{Hypothesis:} The material belongs to the government even after the hand out.
\end{tabular} &
 5819  \\

\midrule
\begin{tabular}[c]{@{}cl@{}} 3 & \textbf{Premise:} 
The emergency department at Harborview probably sees 50 times as many patients with alcohol \\ & problems as the psychiatry or family medicine departments.\\

& \textbf{Hypothesis:} The family medicine departments do not treat alcohol problems.
\end{tabular} &
 2221  \\

\bottomrule
  
\end{tabular}%
}
\caption{Top three helpful and harmful train instances based on IF score and further ranking of them by \dknn for a correctly predicted adversarial text by \NNIF for \Mnli \Alzantot}
\label{tab:example_nnif_word}
\end{table*}

\begin{table*}[t]
\centering
\normalsize
\captionsetup{justification=justified}
\resizebox{\textwidth}{!}{
\begin{tabular}{lc}
\toprule
\multicolumn{2}{l}{\textbf{Original text - label Entailment - prediction Entailment}} \\
\midrule
\multicolumn{2}{l}{\begin{tabular}[c]{@{}l@{}}\textbf{Premise:} Although a seemingly mundane, tactical aspect of business, a firm's inventory strategy reflects its approach to managing risk.\end{tabular}} \\
\multicolumn{2}{l}{\textbf{Hypothesis:} It is possible to determine a firm's risk management philosophy by examining their inventory strategy.} \\

\midrule
\multicolumn{2}{l}{\textbf{\Pruthi - prediction Neutral}} \\
\midrule
\multicolumn{2}{l}{\begin{tabular}[c]{@{}l@{}}\textbf{Premise:} Although a seemingly mundane , \textcolor{red}{tactcial} aspect of business , a firm 's \textcolor{red}{itnventory strategxy} reflects its approach to managing risk.\end{tabular}} \\
\multicolumn{2}{l}{\textbf{Hypothesis:} It is possible to determine a firm 's risk management \textcolor{red}{philkosophy} by examining their inventory \textcolor{red}{strategxy.}} \\

\midrule
\textbf{\textcolor{blue}{Top Helpful}} & \textbf{\textcolor{blue}{NNIF Rank}}\\

\midrule
\begin{tabular}[c]{@{}cl@{}} 1 & \textbf{Premise:} Online investment guru Tokyo Joe was sued by the SEC in a civil fraud case. \\
&  \textbf{Hypothesis:} Tokyo Joe has been sued before.
\end{tabular} &
  5091 \\

\midrule
\begin{tabular}[c]{@{}cl@{}} 2 & \textbf{Premise:} Wesray's purchase of Avis was trendy in three ways. \\
& \textbf{Hypothesis:} There are three reasons why Wesray's purchase of Avis is trendy.
\end{tabular} &
  266 \\

\midrule
\begin{tabular}[c]{@{}cl@{}}3 & \textbf{Premise:} yeah i don't mind that um my husband never cared for fast food so we didn't go that often but you know i have no \\ & problem with uh going to a McDonald's or a Wendy's. \\ 
& \textbf{Hypothesis:}My husband and I did not eat that much fast food.
\end{tabular} &
  5630 \\
  
\midrule
\textbf{\textcolor{orange}{Top Harmful}} & \textbf{\textcolor{orange}{NNIF Rank}}\\
\midrule
\begin{tabular}[c]{@{}cl@{}} 1 & \textbf{Premise:} In accordance with the prescribed statutory process, on August 17, 2001, we reported to the Congress, \\ &the President, the Vice President, and other officials that the NEPDG had not provided the requested records.
\\
& \textbf{Hypothesis:} We told Congress that the NEPDG had failed to give us the records.
\end{tabular} &
 1346 \\

 \midrule
\begin{tabular}[c]{@{}cl@{}} 2 & \textbf{Premise:} The case for not acting until you have to was put most vividly by Senate Assistant Majority Leader Don Nickles\\ & of Oklahoma, in a remark that also captures the hard-nosed attitude regarding humanitarian concerns.
\\
& \textbf{Hypothesis:} The statement summarized their sentiment.
\end{tabular} &
 1044  \\

\midrule
\begin{tabular}[c]{@{}cl@{}} 3 & \textbf{Premise:} Predicting that he would get a lot of heat for treating the minister with respect, Novak said that Farrakhan was more measured \\ &and a lot less confrontational and provocative than a lot of the politicians we talk to regularly on this program.
\\
& \textbf{Hypothesis:} Predicting he would get a lot of hear for respecting the minister, Novak said Farrakhan was measured \\ &and less confrontational.
\end{tabular} &
 5623 \\

\bottomrule
  
\end{tabular}%
}
\caption{Top three helpful and harmful train instances based on IF score and further ranking of them by \dknn for a correctly predicted adversarial text by \NNIF for \Mnli \Pruthi}

\label{tab:example_nnif_char}
\end{table*}

\begin{table*}[t]
\centering
\normalsize
\captionsetup{justification=justified}
\resizebox{\textwidth}{!}{
\begin{tabular}{lc}
\toprule
\multicolumn{2}{l}{\textbf{Original text - label Neutral - prediction Neutral}} \\
\midrule
\multicolumn{2}{l}{\begin{tabular}[c]{@{}l@{}}\textbf{Premise:} And, instead of providing an open-ended guarantee on prices to its distributors, the company would guarantee the price for\\ only two weeks after purchase by the distributor, refusing to take back computers unless they malfunctioned.\end{tabular}} \\
\multicolumn{2}{l}{\textbf{Hypothesis:} The distributor could potentially lose out due to this method.} \\

\midrule
\multicolumn{2}{l}{\textbf{\Pruthi - prediction Contradiction}} \\
\midrule
\multicolumn{2}{l}{\begin{tabular}[c]{@{}l@{}}\textbf{Premise:} And , instead of providing an \textcolor{red}{openedned guaantee} on \textcolor{red}{pices} to its disrtibutors , the \textcolor{red}{comapny wuld guaantee} the price for\\ only two weeks after purchase by the distributor, refusing to take \textcolor{red}{badk} computers \textcolor{red}{ulness} they malfunctioned .\end{tabular}} \\
\multicolumn{2}{l}{\textbf{Hypothesis:} \textcolor{red}{Tehe} distributor could \textcolor{red}{poteIntially} lose out \textcolor{red}{de} to this \textcolor{red}{mehgod} . } \\

\midrule
\textbf{\textcolor{blue}{Top Helpful}} & \textbf{\textcolor{blue}{NNIF Rank}}\\

\midrule
\begin{tabular}[c]{@{}cl@{}} 1 & \textbf{Premise:} No it was gas because you washed your legs all over because you did it in shorts. \\
&  \textbf{Hypothesis:} Your legs were washed all over due to having done it in shorts.
\end{tabular} &
  2373 \\

\midrule
\begin{tabular}[c]{@{}cl@{}} 2 & \textbf{Premise:} Leaving the British official who twice searched his luggage none the wiser, he managed by meticulous observation to memorize the \\ &principal features of the power loom well enough to produce his own version of it on his return to Boston. \\
& \textbf{Hypothesis:} He failed at retaining the information in his head but managed to build a rough prototype of the power loom anyway.
\end{tabular} &
  4706 \\

\midrule
\begin{tabular}[c]{@{}cl@{}}3 & \textbf{Premise:} Bin Ladin shares Qutb's stark view, permitting him and his followers to rationalize even unprovoked mass murder as \\ &righteous defense of an embattled faith. \\ 
& \textbf{Hypothesis:} Bin Ladin views his actions as a defense of his faith.
\end{tabular} &
  925 \\
  
\midrule
\textbf{\textcolor{orange}{Top Harmful}} & \textbf{\textcolor{orange}{NNIF Rank}}\\

\midrule
\begin{tabular}[c]{@{}cl@{}} 1 & \textbf{Premise:} As graduates of the class of 1990, we would like to leave behind something tangible, in appreciation for the support and \\ &encouragement we have received from other students in the School of Engineering and Technology.
\\
& \textbf{Hypothesis:} We want leave a concrete symbol of our appreciation to the school.
\end{tabular} &
 1336 \\

 \midrule
\begin{tabular}[c]{@{}cl@{}} 2 & \textbf{Premise:} Fortunately, not all reports are as disturbing as Hochschild's.
\\
& \textbf{Hypothesis:} Thankfully, not all reports are as terrifying as Hochschild's.
\end{tabular} &
 4817  \\

\midrule
\begin{tabular}[c]{@{}cl@{}} 3 & \textbf{Premise:} Of the two, the W geographical listings seem more  W lists Aylesbury, which, through some grievous, egregious fault, is not \\ &in the geographical section of the L but does appear in the A-Z section (because of the ducks).
\\
& \textbf{Hypothesis:} For some reason, the ducks put the topic in the A-Z section.
\end{tabular} &
 268 \\

\bottomrule
\end{tabular}%
}
\caption{\Mnli \Pruthi adversarial text that the \NNIF fails to detect; showing top three helpful and harmful train instances based on IF score and further ranking of them by \dknn}

\label{tab:example_nnif_incorr_char}
\end{table*}

\begin{table*}
\centering
\normalsize
\captionsetup{justification=justified}
\resizebox{\textwidth}{!}{
\begin{tabular}{lc}
\toprule
    \multicolumn{2}{l}{\textbf{Original text - label Entailment prediction Entailment}}
    \\\midrule
    \multicolumn{2}{l}{\begin{tabular}[c]{@{}l@{}}\textbf{Premise}: Finally, it might be worth mentioning that the program has the capacity to store in a temporary memory buffer about \\
    100 words (proper names, for instance) that it has identified as not stored in its dictionary. \end{tabular}} \\
    \multicolumn{2}{l}{\textbf{Hypothesis}: It's possible to store words in a temporary dictionary, if they don't appear in a regular dictionary.} \\
    \midrule
    \multicolumn{2}{l}{\textbf{\Alzantot - prediction Neutral}}
    \\
    \midrule
    \multicolumn{2}{l}{\begin{tabular}[c]{l}\textbf{Premise}: Finally, it might be worth \textcolor{red}{mentioning} that the program has the capacity to store in a temporary memory \textcolor{red}{buffer} about \\
    100 words (proper names, for instance) that it has identified as not stored in its dictionary.\end{tabular}}\\
    
    \multicolumn{2}{l}{\textbf{Hypothesis}: It's possible to \textcolor{red}{shopping} words in a temporary dictionary, if they don't appear in a regular dictionary. }\\
    
    \midrule
    \multicolumn{2}{l}{\textbf{Repaired text by \DISP}}\\
    \midrule
    \multicolumn{2}{l}{\begin{tabular}[c]{l}\textbf{Premise}: Finally, it might be worth \textcolor{blue}{that} that the program has the capacity to store in a temporary memory buffer about \\
    100 words (proper names, for instance) that it has identified as not stored in its dictionary.\end{tabular} }\\
    \multicolumn{2}{l}{\textbf{Hypothesis}: It's possible to \textbf{\textcolor{blue}{do}} words in a temporary dictionary, if they don't appear in a regular dictionary.}\\
    
    \midrule
    \multicolumn{2}{l}{\textbf{Repaired text by \FGWS }}\\
    \midrule
    \multicolumn{2}{l}{\begin{tabular}[c]{l}\textbf{Premise}: Finally, it might be worth \textbf{\textcolor{blue}{name}} that the program has the capacity to store in a temporary memory \textcolor{blue}{pilot} about \\
    100 words (proper names, for instance) that it has identified as not stored in its dictionary.\end{tabular} }\\
    \multicolumn{2}{l}{\textbf{Hypothesis}: It's possible to \textcolor{blue}{shopping} words in a temporary dictionary, if they don't appear in a regular dictionary.} \\

    \midrule
    \textcolor{blue}{\textbf{Top Helpful}} & \textcolor{blue}{\textbf{NNIF Rank}}\\

    \midrule
    \begin{tabular}[c]{@{}cl@{}} 1 & \textbf{Premise:} yeah i mean they're they're throwing more money at it now than ever before and things are getting worse.\\
& \textbf{Hypothesis:} The money is going to the wrong things, so it's not fixing the problem.
\end{tabular} &
 1295 \\

\midrule
\begin{tabular}[c]{@{}cl@{}} 2 & \textbf{Premise:}I put \$75 on the New England Patriots as a 2.5-point underdog and \$50 on a Boston Red Sox playoff \\ & game against the Cleveland Indians.\\
& \textbf{Hypothesis:} I bet a total of \$125 dollars on the New England Patriots and the Boston Red Sox. 
\end{tabular} &
 3870  \\

\midrule
\begin{tabular}[c]{@{}cl@{}} 3 & \textbf{Premise:} Graffiti written by Russian soldiers can be seen in the caves of Antiparos.\\
& \textbf{Hypothesis:} Russian soldiers drew graffiti on the walls of The Louvre.
\end{tabular} &
 4122  \\
    \midrule
    \textbf{\textcolor{orange}{Top Harmful}} & \textbf{\textcolor{orange}{NNIF Rank}}\\   
    \midrule
    \begin{tabular}[c]{@{}cl@{}} 1 & \textbf{Premise:} But I am most serious. \\
& \textbf{Hypothesis:} I'm not joking at all.
\end{tabular} &
 808 \\

\midrule
\begin{tabular}[c]{@{}cl@{}} 2 & \textbf{Premise:} the uniformed services, recognize that promotional material received by a uniformed service member \\ & traveling on official business at government expense belongs to the government and must be relinquished in \\ & accordance with service regulations\\
& \textbf{Hypothesis:} TThe material belongs to the government even after the hand out.
\end{tabular} &
 2835  \\

\midrule
\begin{tabular}[c]{@{}cl@{}} 3 & \textbf{Premise:} According to NIST, accreditation is the formal authorization by the management \\ &official for system operation and an explicit acceptance of risk.\\
& \textbf{Hypothesis:} Accreditation is the formal authorization by the management official for system operation \\ & and an explicit acceptance of risk, according to NIST.
\end{tabular} &
 2908  \\
    \bottomrule
    \end{tabular}%
    }
    \caption{Example sentence from \citet{liu-etal-2022-detecting} that is predicted correctly by \NNIF, \DISP, \LID, \MDRE and incorrectly by \FGWS }
    
\label{tab:example_nnif_disp}
\vspace{-1cm}
\end{table*}

\end{document}